\documentclass{ecai} 
\usepackage{latexsym}
\usepackage{amssymb}
\usepackage{amsmath}
\usepackage{amsthm}
\usepackage{booktabs}
\usepackage{enumitem}
\usepackage{graphicx}
\usepackage{subfig} 
\usepackage{color}
\usepackage{algorithm}
\usepackage{pifont}
\usepackage{url}
\usepackage{amsmath,amssymb}
\usepackage{algpseudocode}

\newcommand{\BibTeX}{B\kern-.05em{\sc i\kern-.025em b}\kern-.08em\TeX}

\begin{document}

%%%%%%%%%%%%%%%%%%%%%%%%%%%%%%%%%%%%%%%%%%%%%%%%%%%%%%%%%%%%%%%%%%%%%%%%

\begin{frontmatter}

%%% Use this command to specify your submission number.
%%% In doubleblind mode, it will be printed on the first page.

\paperid{6754} 

%%% Use this command to specify the title of your paper.

\title{Leveraging Latent Causal Relationships Among Web Services for Traffic Prediction}

%%% Use this combinations of commands to specify all authors of your 
%%% paper. Use \fnms{} and \snm{} to indicate everyone's first names 
%%% and surname. This will help the publisher with indexing the 
%%% proceedings. Please use a reasonable approximation in case your 
%%% name does not neatly split into "first names" and "surname".
%%% Specifying your ORCID digital identifier is optional. 
%%% Use the \thanks{} command to indicate one or more corresponding 
%%% authors and their email address(es). If so desired, you can specify
%%% author contributions using the \footnote{} command.

\author[A]{\fnms{Chang}~\snm{Tian}}
\author[B]{\fnms{Mingzhe}~\snm{Xing}\footnote{Co-first author.}\thanks{Mingzhe Xing and Yinliang Yue are corresponding authors.}}
\author[C]{\fnms{Zenglin}~\snm{Shi}}
\author[A]{\fnms{Matthew}~\snm{Blaschko}}
\author[B]{\fnms{Yinliang}~\snm{Yue}}
\author[A]{\fnms{Marie-Francine}~\snm{Moens}}

% \author[B,C]{\fnms{Third}~\snm{Author}\orcid{....-....-....-....}} 

\address[A]{KU Leuven}
\address[B]{Beijing Zhongguancun Laboratory}
\address[C]{Hefei University of Technology}

\address{chang.tian@kuleuven.be,xingmz@zgclab.edu.cn}
% \address[B]{Short Affiliation of Second Author and Third Author}
% \address[C]{Short Alternate Affiliation of Third Author}

%%% Use this environment to include an abstract of your paper.

\begin{abstract}
Predicting web service traffic is crucial for system operation tasks including dynamic resource scaling, anomaly detection, and fraud detection. 
Web service traffic is characterized by frequent and drastic fluctuations over time and are influenced by heterogeneous user behaviors, making accurate prediction a challenging task. 
Previous research has extensively explored statistical approaches, and neural networks to mine features from preceding service traffic time series for prediction. 
However, these methods have largely overlooked the latent causal relationships between services. 
Drawing inspiration from causality in ecological systems, we empirically recognize the causal relationships between web services. 
To leverage these relationships for improved traffic prediction, we propose an effective neural network module, CCMPlus, designed to extract causal relationship features across services. 
This module can be seamlessly integrated with existing time series models to consistently enhance the performance of traffic predictions. 
We theoretically justify that the causal correlation matrix generated by the CCMPlus module captures causal relationships among services.
Empirical results on real-world datasets from Microsoft Azure, Alibaba Group, and Ant Group confirm that our method surpasses state-of-the-art approaches in Mean Squared Error and Mean Absolute Error for predicting service traffic time series. 
These findings highlight the efficacy of feature representations from the CCMPlus module.
\end{abstract}

\end{frontmatter}

%%%%%%%%%%%%%%%%%%%%%%%%%%%%%%%%%%%%%%%%%%%%%%%%%%%%%%%%%%%%%%%%%%%%%%%%

\section{Introduction}
% Artificial intelligence (AI) techniques power a diverse range of applications, including computer vision~\cite{liu2024visual, li2021paint4poem, liu2024improved} and natural language processing~\cite{tian2022anti,touvron2023llama,tian-etal-2024-generic, kalla2023study}, both of which play a pivotal role in supporting various social and practical activities. As AI continues to advance, the use of web services has grown substantially~\cite{shen2024artificial, martin2020ai}. Predicting web service traffic carries significant social and practical value, with applications spanning dynamic resource scaling~\cite{pan2023magicscaler,zou2024optscaler}, load balancing~\cite{pavlenko2024vertically}, system anomaly detection~\cite{mitropoulou2024anomaly}, service-level agreement compliance~\cite{liao2024retrospecting}, fraud detection~\cite{Al-talak2021}, and so on. These capabilities not only enhance system performance but also improve the overall user experience with these technologies~\cite{kumar2024tpmcf}.

User-oriented web services continue to grow exponentially, which has significantly accelerated the development of a diverse range of customized web applications~\citep{li2021paint4poem, tian2022anti,tian2024fighting, tian2024generic,tian2025large}. 
These services attract a substantial user base and play a pivotal role in enabling various social and practical activities. 
For instance, YouTube has amassed 2.7 billion users\footnote{https://www.globalmediainsight.com/blog/youtube-users-statistics/}, powered by its cutting-edge recommendation algorithms.
Accurately predicting web service traffic carries significant social and practical value, with applications spanning dynamic resource scaling~\cite{pan2023magicscaler,zou2024optscaler}, load balancing~\cite{pavlenko2024vertically}, anomaly detection~\cite{mitropoulou2024anomaly}, service-level agreement compliance~\cite{liao2024retrospecting}, fraud detection~\cite{Al-talak2021}, and so on. 
These capabilities not only enhance system performance but also improve the overall user experience~\cite{kumar2024tpmcf}.

Co-located, long-running web services often experience diverse workload patterns~\cite{zou2024optscaler}. 
Web service traffic is characterized by frequent and significant fluctuations over time, driven by heterogeneous user behaviors. 
These factors collectively make predicting web service traffic a highly challenging task~\cite{straesser2025trust, pan2023magicscaler}.
Previous works formulates this task as a typical time series forecasting task. 
These approaches can broadly be categorized into statistical~\cite{kapgate2014weighted, hu2016autoscaling, kumar2016forecasting}, machine learning~\cite{issa2017using, daraghmeh2018linear, hu2019stream}, and deep learning~\cite{ruan2023workload, guo2018applying, pan2023magicscaler, zou2024optscaler} methods.
While statistical methods struggle to handle multi-dimensional and non-linear traffic data, machine learning methods address these limitations but fail to achieve the same level of accuracy as deep learning approaches~\cite{zou2024optscaler, alharthi2024auto}. 
The recent advancements in Transformer~\cite{vaswani2017attention} architecture have demonstrated superior performance in sequential prediction tasks. 
Consequently, Transformer-based methods~\cite{qi2022performer} have emerged, achieving promising results in web service traffic prediction.
Aside from these ad-hoc models, general state-of-the-art (SOTA) Transformer-based time-series prediction methods~\cite{zhou2022fedformer, liuitransformer, chenpathformer, wang2024timexer} also have the potential to be applied to the web traffic task.
Recently emerged LLM-based methods~\cite{tan2024language, touvron2023llama, jintime} leverage advanced reasoning capabilities by processing time series data through specially designed prompts, offering a promising avenue for temporal modeling.
% However, they did not pay enough attention to the causal relationship across web services.

\begin{figure}[tbp]
  \centering
  \includegraphics[width=0.99\columnwidth]{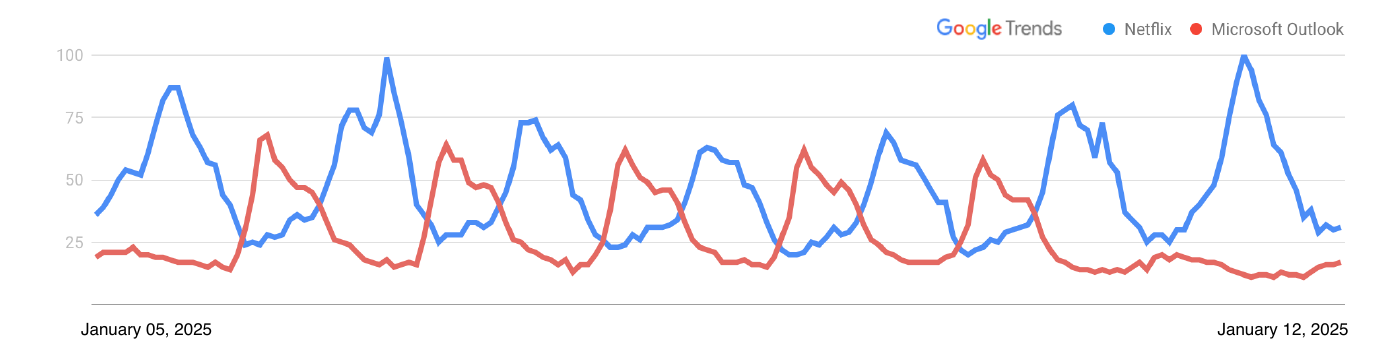}
  \caption{This figure illustrates the 7-day normalized search interest data for Netflix and Microsoft Outlook from Google Trends in the UK. The two services exhibit a causal relationship: an increase in Netflix usage corresponds to a decrease in the use of the email service.}
  \vspace{0.5cm}
  \label{fig:services_causality}
\end{figure}

% \begin{figure}[t]
%   \centering
%     \includegraphics[width=0.7\columnwidth]{figures/ccm_intro.pdf}
%   \caption{Illustration of ecological and analogical web service causality. The rabbit population influences grass abundance, where an increase in rabbits leads to a decrease in grass. This principle inspires our CCMPlus module, which captures causal effects among web services.}
%   \label{fig:rabbit}
% \end{figure}

Drawing inspiration from ecological causality, for example, grass abundance and rabbit populations influencing one another iteratively~\cite{ji2022superprocesses}, we identify analogous patterns in web service traffic as shown in Figure~\ref{fig:services_causality}. 
Empirical observations of Google Trends data reveal a causal relationship between leisure websites, such as Netflix, and work-related software, like Outlook. 
Specifically, increased web traffic to Netflix corresponds to decreased traffic to Outlook, and vice versa. 
These empirical observations suggest the existence of latent causal relationships underlying human-driven web behaviors. 
By uncovering and leveraging these causal relationships, we can achieve more accurate modeling of service traffic patterns.
Building on this motivation and the causality theory of Convergent Cross Mapping (CCM)~\cite{sugihara2012detecting}, which originates from ecology, we introduce the CCMPlus module. 
This module extracts features from web service traffic time series while including causal relationships among services, thereby enhancing the accuracy of web service traffic prediction. 
Furthermore, the CCMPlus module could integrate easily with existing time series forecasting models, enriching them with more informative, causally-aware features.

Concretely, we extend the CCM theory to effectively merge with neural networks, resulting in the development of the CCMPlus module. 
Specifically, the traditional CCM theory relies on a set of expert designed hyper-parameters, which might introduce severe bias to the causality identification.
To alleviate this, we propose a multi-manifold embedding space which is constructed by learnable parameters and present the causality relationship in diverse spaces.
After obtaining the intial embedding, we employ the CCM procedure but in the multi-manifold space to calculate the causal correlation matrix, which is updated with a momentum update mechanism.
By integrating the causal correlation matrix to the initial embedding, it generates a resulting feature representation that incorporates informative causal information. 
This enhanced feature representation can then be concatenated with the feature representations of web service traffic time series extracted by existing time series models, thereby improving prediction accuracy.
Our main contributions\footnote{The code, data, and other resources will be available at https://github.com/changtianluckyforever/CCMPlus.} in this work can be summarized as follows:
\begin{itemize}
    \item \textbf{Method:} 
    The CCMPlus module enhances existing time series forecasting models by generating feature representations that incorporate latent causal relationships across web services, addressing a critical limitation of many previous methods. Additionally, the CCMPlus module is designed for seamless integration with existing time series forecasting models, further contributing to improved prediction accuracy.
    \item \textbf{Theory:} We justify that the causal correlation matrix generated by the CCMPlus module effectively captures causal relationships across web services, enabling the incorporation of these relationships into web traffic prediction methods.
    \item \textbf{Experiments:} Experiments conducted on three real-world web service traffic datasets (Alibaba Group, Microsoft Azure, and Ant Group) demonstrate that our method achieves superior performance in terms of Mean Squared Error (MSE) and Mean Absolute Error (MAE) compared to previous state-of-the-art methods, thereby validating the effectiveness of the CCMPlus module.
\end{itemize}

% The remainder of the paper is organized as follows. Section 2 reviews related work. Section 3 provides preliminary information about our proposed method. Section 4 elaborates on the proposed framework, encompassing the CCMPlus and time series backbone modules. Section 5 compares experimental results from the proposed method and prevalent methods, and analyzes the method and results. Section 6 concludes this paper.

\section{Related Work}
\label{sec:rlwk}
\subsection{Web Service Traffic Prediction}
Web service traffic prediction is a task critical to enabling service autoscaling, load balancing, and anomaly detection. As web service traffic is often represented as time series data, existing approaches primarily frame traffic prediction as a time series prediction problem~\cite{pan2023magicscaler,zou2024optscaler,alharthi2024auto}.

Early research focused on statistical prediction methods~\cite{kapgate2014weighted, hu2016autoscaling, kumar2016forecasting} such as Moving Average, Auto-Regression, and Autoregressive Integrated Moving Average. These methods are valued for their simplicity and interpretability but are constrained by strict stationarity requirements. Moreover, they struggle to extend to multi-dimensional or non-linear data, limiting their applicability in dynamic environments.
To overcome these limitations, methods like HOPBLR~\cite{issa2017using}, LLR~\cite{daraghmeh2018linear} and TWRES~\cite{hu2019stream} employed machine learning techniques, including Logistic Regression and Support Vector Regression, respectively, for traffic prediction. However, these approaches were hindered by the limited expressive capacity of their models, resulting in suboptimal prediction accuracy.
More recent advancements have shifted towards deep learning methods. CrystalLP~\cite{ruan2023workload} and GRUWP~\cite{guo2018applying} utilize Long Short-Term Memory networks and Gated Recurrent Units, respectively, to predict service workloads. 
% L-PAW~\cite{} introduces a top-sparse autoencoder to derive sparse representations from highly dynamic service data. 
MagicScaler~\cite{pan2023magicscaler} proposes a novel multi-scale attentive Gaussian process-based predictor, capable of accurately forecasting future demands by capturing scale-sensitive temporal dependencies. 
% It employs a two-stage feature extraction process and Gaussian processes to predict uncertain future demands with high precision.
The Performer~\cite{qi2022performer} integrates the Transformer architecture into an encoder-decoder paradigm for service workload prediction. It leverages the self-attention mechanism to model temporal correlations and learn both global and local representations effectively.
OptScaler~\cite{zou2024optscaler} advances this direction with a proactive prediction module comprising a long-term periodic block and a short-term local block to capture multi-scale temporal dependencies. 
While existing traffic prediction methods demonstrate high accuracy through advanced time series forecasting techniques, they largely neglect the underlying causal relationships within the web services. Exploring these hidden causalities holds significant potential for further improving prediction performance.

\subsection{Time Series Forecasting}
Besides approaches specifically tailored for web service traffic prediction, general time series forecasting methods are also applied to predict web traffic~\cite{alharthi2024auto,zou2024optscaler,10457027}.

Traditional statistical methods such as Prophet~\cite{sean2018forecasting}, and Holt-Winters~\cite{hyndman2018forecasting} assume that time series variations adhere to predefined patterns. However, the inherently complex fluctuations of web service traffic often exceed the scope of these predefined patterns, thereby limiting the practical applicability of such statistical methods~\cite{wutimesnet}.

Recent advancements in neural network architectures have significantly enhanced temporal modeling capabilities. Neural network approaches for time series forecasting can be categorized into five paradigms~\cite{wangtimemixer, tan2024language}: RNN-based, CNN-based, Transformer-based, MLP-based, and large language model (LLM)-based methods. 
% CNN-based methods~\cite{wang2023micn, hewage2020temporal} use convolutional kernels along the temporal dimension to capture sequential patterns, while RNN-based methods~\cite{franceschi2019unsupervised,dudukcu2023temporal,peng2024reservoir} employ recurrent structures to model temporal state transitions. Transformer-based methods~\cite{zhou2022fedformer,liuitransformer,chenpathformer,wang2024timexer} are widely recognized for their effective feature extraction through attention mechanisms. LLM-based methods~\cite{tan2024language,touvron2023llama,jintime} leverage advanced reasoning capabilities, presenting a promising avenue for temporal modeling. Additionally, MLP-based methods~\cite{wang2024timexer,olivares2023neural,daslong,wutimesnet} offer a compelling balance of forecasting accuracy and computational efficiency.
Empirical methods often integrate components from the aforementioned categories, utilizing specific designs to effectively capture critical temporal features~\cite{wangtimemixer}. These specific designs incorporate series decomposition, multi-periodicity analysis, and multi-scale mixing architectures.
For series decomposition, Autoformer~\cite{wu2021autoformer} introduces a decomposition block based on moving averages, enabling the separation of complex temporal variations into seasonal and trend components. Building on this foundation, DLinear~\cite{zeng2023transformers} utilizes series decomposition as a preprocessing step prior to performing linear regression. Crossformer~\cite{zhang2023crossformer} segments time series data into subseries-level patches and employs a Two-Stage Attention layer to effectively model cross-time and cross-variable dependencies within each patch. iTransformer~\cite{liuitransformer} leverages the global representation of entire series and applies attention mechanisms to these series-wise representations, facilitating the capture of multivariate correlations. TimeXer~\cite{wang2024timexer} integrates external information into the Transformer architecture through a carefully designed embedding strategy, allowing the inclusion of external information into patch-wise representations of endogenous series. In the context of multi-periodicity, NBEATS~\cite{oreshkinn} employs multiple trigonometric basis functions to model time series, providing a robust framework for handling periodic patterns. Similarly, TimesNet~\cite{wutimesnet} applies Fourier Transform to decompose time series into components of varying periodic lengths and utilizes a modular architecture to process these decomposed components effectively. With respect to multi-scale mixing architectures, Pathformer~\cite{chenpathformer} adopts multi-scale patch representations and applies dual attention mechanisms across these patches to capture both global correlations and local details, thereby addressing temporal dependencies comprehensively. TimeMixer~\cite{wangtimemixer} captures temporal features by introducing a novel multi-scale mixing architecture, which comprises two key components: Past-Decomposable-Mixing, designed to leverage disentangled series for multi-scale representation learning, and Future-Multipredictor-Mixing, which ensembles complementary forecasting skills across multi-scale series to enhance prediction accuracy.

While general time series forecasting methods have been applied to web service traffic prediction~\cite{pan2023magicscaler, zou2024optscaler,catillo2023survey}, these methods often overlook causal relationships between services. In contrast, our CCMPlus module computes a causal correlation matrix used to generate temporal features incorporating causal relationships, significantly improving web service traffic prediction accuracy.

\section{Background}
In this section, we begin by giving a formal definition of web service traffic prediction task. 
Subsequently, we provide an overview of the causality theory, Convergent Cross Mapping (CCM), which serves as the theoretical foundation for our proposed CCMPlus module.

\subsection{Web Service Traffic}
\begin{figure}[tp]
  \centering
    \includegraphics[width=0.95\columnwidth]{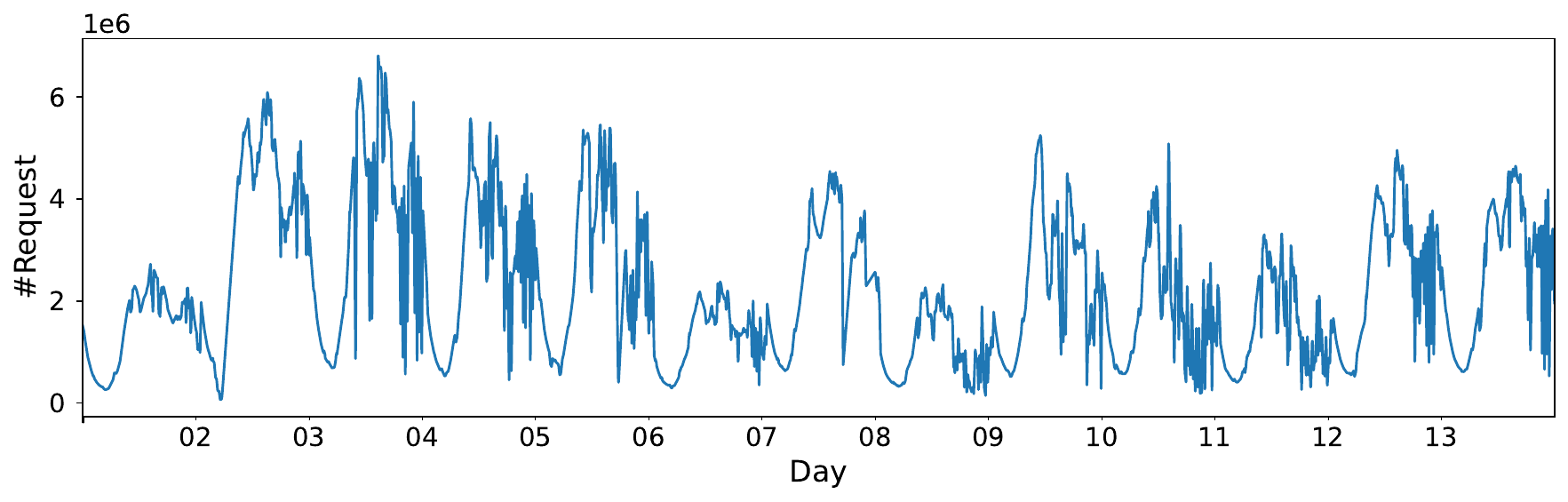}
  \caption{Web service traffic time series from Alibaba Cloud. Each point indicates the number of service requests at each time point.}
  \label{fig:a_azure_web_service}
  \vspace{0.5cm}
\end{figure}
Figure~\ref{fig:a_azure_web_service} presents an illustrative example of web service traffic time series from Alibaba Cloud.
It exhibits significant fluctuations and frequent changes, primarily driven by human behavior and activity patterns.
Accurately predicting web service traffic remains a well-recognized challenge due to the inherent complexity of traffic patterns, as highlighted by the research community~\cite{pan2023magicscaler, zou2024optscaler}.
The web traffic prediction task can be formulated as follows:
\begin{equation}
y(t) = P(y(t-\alpha), y(t-2\alpha), \dots, y(t-k\alpha)), \label{eq:define}
\end{equation}
where $y(t)$ is the number of request at time $t$, $\alpha$ denotes the prediction granularity, and $k$ is the historical sequence length.

\subsection{Convergent Cross Mapping}
\textbf{Convergent Cross Mapping} (CCM) theory published in Science~\cite{sugihara2012detecting} was originally proposed in the field of ecology and is designed to detect causal relationships between species.
The original work includes around 50 pages of supplementary material explaining the CCM theory.
The exposition of CCM is often illustrated using the context of the Lorenz system.
Its trajectory forms a manifold \( M \) in the state space, which consists of a collection of points that represent all possible states  over time, with these points connected to create a structured geometric space. 

The shadow manifolds, \( M_x \) or \( M_y \), represents the projection of the original manifold \( M \) onto the system variables \( X \) or \( Y \), respectively. Specifically, a lagged coordinate embedding utilizes \( E \) time-lagged values of \( X(t) \) as coordinate axes to reconstruct the shadow manifold \( M_x \). A point on \( M_x \), denoted as \( \underline{x}(t) \), is an \( E \)-dimensional vector expressed as:
\begin{equation}
\underline{x}(t) = [X(t), X(t-\tau), X(t-2\tau), \dots, X(t-(E-1)\tau)],
\nonumber
\end{equation}
where \( \tau \) is a positive time lag, and \( E \) denotes the embedding dimension.
Similarly, the same approach 
can be applied to
points \( \underline{y}(t) \) in the shadow manifold \( M_y \).

\begin{figure}[tbp]
  \centering
  \includegraphics[width=0.99\linewidth,keepaspectratio]{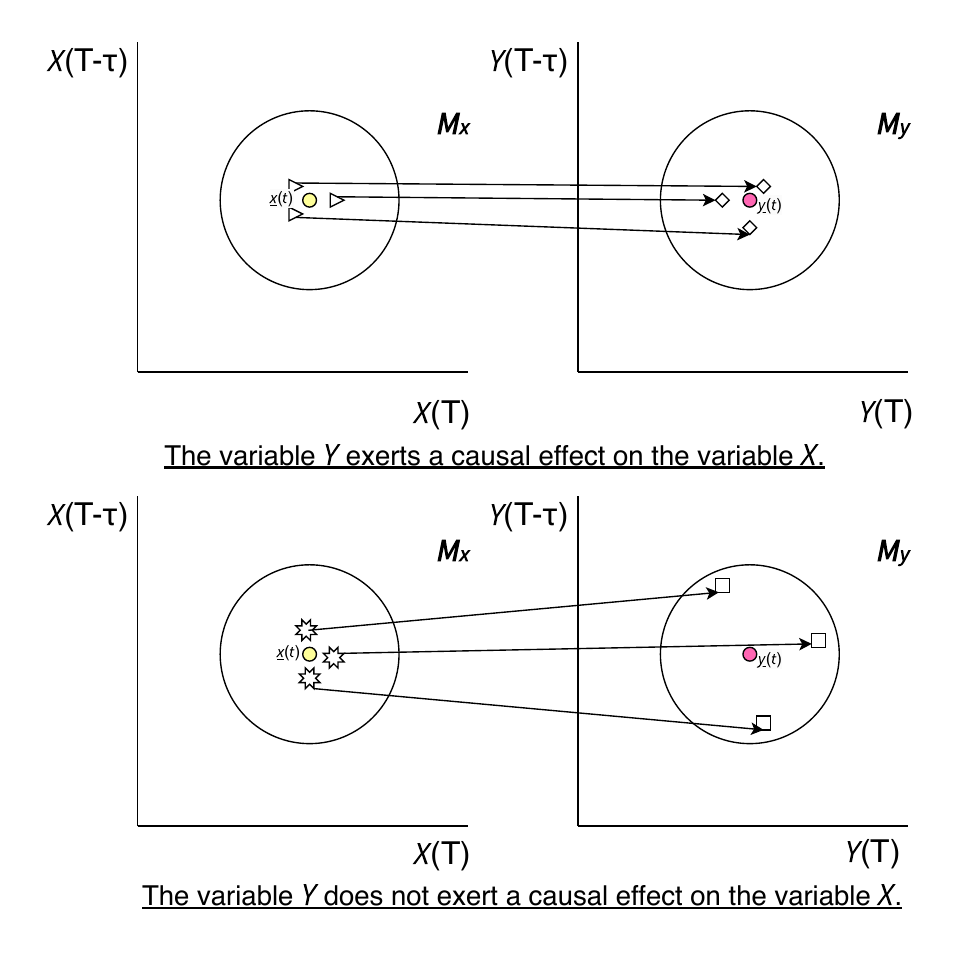} 
  \caption{The point \( \underline{y}(t) \) in the manifold \( M_y \) corresponds to the contemporaneous point in time \( \underline{x}(t) \) in the manifold \( M_x \).}
  \vspace{0.5cm}
  \label{fig:ccm_cross_mapping_main}
\end{figure}

\textbf{Cross mapping} refers to the process of identifying contemporaneous points in the manifold \( M_x \) of one variable \( X \) based on points in the manifold \( M_y \) of another variable \( Y \). 
Specifically, given a point \( \underline{y}(t) \) in the manifold \( M_y \), the corresponding point in time from the manifold \( M_x \) is \( \underline{x}(t) \). 
As illustrated in Figure~\ref{fig:ccm_cross_mapping_main}, if \( Y \) exerts a causal effect on \( X \), information from \( Y \) will be stored in \( X \). Consequently, the neighbors of \( \underline{x}(t) \) in \( M_x \) will correspond to points with the same time indices in \( M_y \), and these corresponding points will also be neighbors of \( \underline{y}(t) \). However, if \( Y \) has no causal effect on \( X \), the information about \( Y \) in \( X \) will be incomplete~\cite{sugihara2012detecting}. 
As a result, the timely corresponding points in \( M_y \) will diverge and no longer be neighbors of \( \underline{y}(t) \).

% \begin{figure}[t]
%   \centering
%   \includegraphics [width=\columnwidth]{figures/new_lorenz_system_compressed.pdf}
%   \caption{The Convergent Cross Mapping (CCM) theory is often explained using the Lorenz system, utilizing its original manifold and corresponding shadow manifolds.}
%   \label{fig:ccm_lorenz_system}
% \end{figure}

% \begin{figure}[tbp]
%   \centering
%   \includegraphics[width=0.85\linewidth,keepaspectratio]{figures/correlation_coefficient_plot.png} 
%   \caption{Convergent predictability as the time series length increases, assuming $Y$ has a causal effect on $X$.
% }
%   \label{fig:ccm_convergence}
% \end{figure}

\textbf{Convergence} in CCM implies that if the variable \( Y \) has a causal effect on \( X \), extending the observation period improves the ability to predict \( Y \) using points on the shadow manifold \( M_x \). 
A longer observation period provides more trajectories to fill the gaps in the manifold, resulting in a more defined structure, which enhances the prediction of \( \hat{Y}(t) \mid M_x \). 
Conversely, if two variables do not have a causal relationship, refining their manifolds will not lead to an improvement in predictive accuracy. 
More details of the CCM algorithm can be found in Appendix Section~\ref{sec:ccm_procedure}.

\section{Methodology}
\begin{figure*}[t]
    \centering
\includegraphics[width=0.99\textwidth,keepaspectratio]{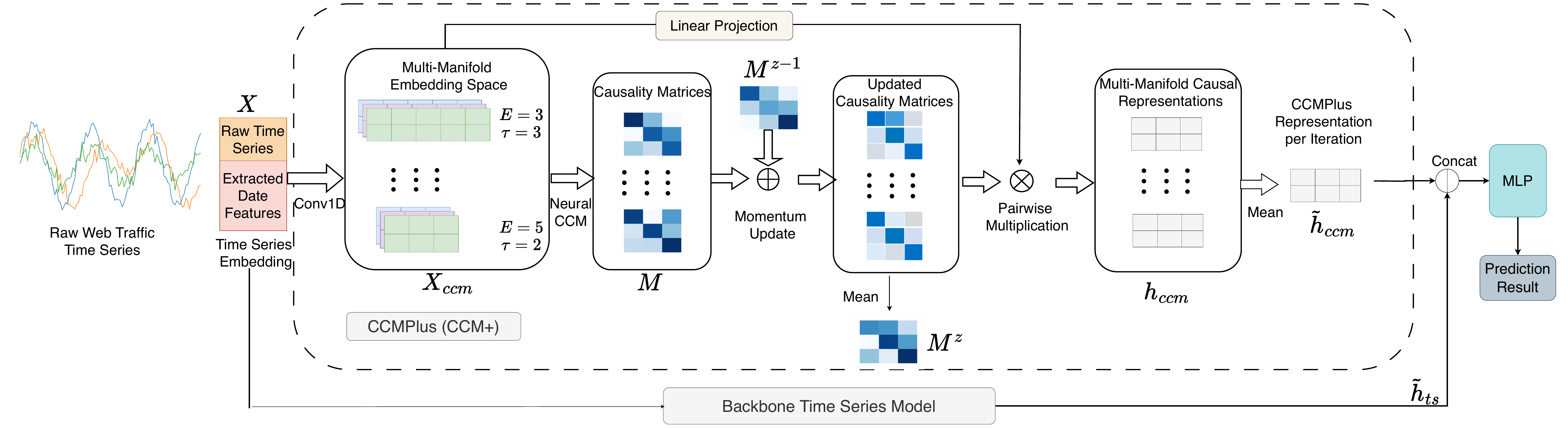}
    \caption{An overview of the proposed method, which consists of the CCMPlus module and the Backbone Time Series Model. The CCMPlus module identifies latent causal relationships among web services, generating a CCMPlus representation that enhances the predictive performance of the Backbone Time Series Model.}
    \label{fig:model}
\end{figure*}

Inspired by the concept of causal relationships in ecology, for example, the population dynamics of rabbits influence the abundance of grass, we analogically observe causal relationships in the web traffic patterns of different web services.
As depicted in Figure~\ref{fig:services_causality}, real-world data from Google Trends indicates that an increase in web traffic for Netflix corresponds to a decrease in traffic for the office software Outlook.
To leverage the latent causality between services for traffic prediction, we extend the CCM theory by integrating it into a neural network module, referred to as CCMPlus (CCM+). 
It leverages causal relationships among web services to generate the CCMPlus representation~(Section~\ref{sec:ccmplus_module}). This representation is used to enhance the representation produced by the Backbone Time Series Model~(Section~\ref{sec:backbone_ts_model}), boosting the accuracy of web service traffic time series prediction.
Finally, we introduce the optimization process in Section~\ref{sec:optimize}.
% denoted as \( \hat{t} \). The predictions (\( \hat{t} \)) are compared against the ground truth (\( t \)) using the mean squared error (MSE) loss, which serves as the optimization objective for the entire model.

% In Section~\ref{sec:ccmplus_module}, we detail the construction of the causal correlation matrix by estimating one web service time series using points from the shadow manifold of another web service time series, and then generate the CCMPlus representation using the causal correlation matrix. Additionally, we briefly introduce the two backbone time series models in Section~\ref{sec:backbone_ts_model}. The pseudocode of CCMPlus module is in Algorithm~\ref{alg:ccmplus_module}.
% Finally, we introduce the optimization process in section~\ref{sec:optimize}.

\subsection{CCMPlus Module}
\label{sec:ccmplus_module}
% first we write   \tau and E how come out?  introduce \tau_w       x and target  x.shape    target.shape   
% convolution  then we have conv_out coordinates
% consistent target_ccm length 
% In this section, we aim to derive the feature representation incorporating the causal relationships across web services time series from raw time series, so as to enhance the prediction precision.
In this section, we aim to derive a feature representation that captures causal relationships across web services from raw time series data, thereby improving the precision of traffic prediction.
This process follows the original CCM algorithm, and constructs the causal correlation matrix by estimating each time point of one web service time series using points from the shadow manifolds of another web service time series.
After that, the casualty matrix is used to generate the CCMPlus representation.

% We first extract date features, e.g., minutes, hours and days, and then convert the date features into embeddings and then combine the embeddinngs with the raw time
% The ground truth time series is denoted as $ target \in \mathbb{R}^{ N\times L_x}$.
% Note that for constructing the shadow manifold, the order of $\hat{\mathbf{X}}$ along the \( L_x \) dimension is reversed. 

\paragraph{\ding{172} Multi-Manifold Embedding}
The first step is to derive the initial time series embedding.
Given the raw time series $\hat{\mathbf{X}} \in \mathbb{R}^{N\times L}$ of length $L$ for $N$ web service traffic , 
% note that for constructing the shadow manifold, the order of $\hat{\mathbf{X}}$ along the \( L \) dimension is reversed.
we first extract date features, such as minutes, hours, and days, convert them into embeddings, and then combine these embeddings with the raw time series $\hat{\mathbf{X}}$ to form the $C$-dimensional time series embedding $\mathbf{X} \in \mathbb{R}^{N\times L\times C}$. 
Next, we follow the original CCM algorithm to transform the time series into shadow manifold space.
Given that the time index of the time‑evolving variable \(X(t)\) ranges from \(1\) to \(L\), the shadow manifold \(M_x\) is constructed by forming lagged coordinate vectors:
% Considering the time index of time-evolving variable \( X(t) \) ranges from \( 1 \) to \( L \), the shadow manifold \( M_x \) is constructed by forming lagged coordinate vectors:
\begin{equation}
        \underline{x}(t) = [X(t), X(t-\tau), X(t-2\tau), \dots, X(t-(E-1)\tau)],\label{eq:manifold_embed}
\end{equation}
where $t\in [1+(E-1)\tau, L]$, \( \tau \) represents the time lag, and \( E \) denotes the embedding dimension.
% Note that for constructing the shadow manifold, the order of $\mathbf{X}$ along the \( L \) dimension is reversed.

The traditional CCM algorithm relies on experts to set the value of $\tau$, which might introduce human bias.
However, since different web time series have different characteristics, determining $\tau$ based on expert knowledge is subjective, labor-intensive and challenging.
To alleviate this issue, we extend it to a \textbf{multi-manifold space}, so as to learn the causality from \textbf{diverse shadow manifold spaces}.
Specifically, we initialize two vectors of \( \mathbf{\tau} = [\tau_1, \ldots, \tau_i, \ldots, \tau_n] \) and the corresponding \( \mathbf{E} = [E_1, \ldots, E_i, \ldots, E_n] \).
For simplicity, we use the \( i \)-th shadow manifold, defined by the pair \( (\tau_i, E_i) \), as an example to explain the CCMPlus procedures. 
The initialized time lag $\tau_i$ and the corresponding embedding dimension $E_i$ are generated as follows:
{\small
\begin{equation}
E_i = 
\begin{cases} 
\left\lfloor \frac{\tau_w}{\tau_i} \right\rfloor - 1, & \text{if } \left\lfloor \frac{\tau_w}{\tau_i} \right\rfloor \bmod 2 = 0, \\[0.5em]
\left\lfloor \frac{\tau_w}{\tau_i} \right\rfloor, & \text{if } \left\lfloor \frac{\tau_w}{\tau_i} \right\rfloor \bmod 2 = 1,\nonumber
\end{cases} 
% \quad \forall \tau_i \in \tau_{\text{list}},\nonumber
\end{equation}
}
where \( \tau_w \) is set to $100$ based on prior empirical observations for shadow manifolds as established in the literature~\cite{kugiumtzis1996state}.
Given the pair \( (\tau_i, E_i) \), the corresponding calculation process of manifold embedding in Eq.~\ref{eq:manifold_embed} can be viewed as a convolution process, and can be effectively computed with the convolution neural network.
Therefore, the convolution output can be derived as follows:
\begin{equation}
\mathbf{X}_{conv} = \text{Conv1D}\bigl({\mathbf{X}};\, \text{kernel size}=E_i, \text{dilation}=\tau_i\bigr),\nonumber
\end{equation}
where ${\mathbf{X}}$ is reversed along the \( L \) dimension, 
the channels of input and output of the convolution network are $C$ and $D$, respectively. 
The trajectory length of the \( i \)-th shadow manifold can be derived as $\bar{L} = L - \tau_i (E_i - 1)$.

After that, we can derive the shadow manifold embedding $\mathbf{X}_{ccm} \in \mathbb{R}^{N\times \bar{L}\times D}$ reshaped from $\mathbf{X}_{conv}$, where, for each time point in \( \bar{L} \), the corresponding point in the shadow manifold is represented by coordinates in \( D \) dimensions. 
Specifically, \( P = \mathbf{X}_{ccm}[m, :, :] \), with \( P \in \mathbb{R}^{\bar{L} \times D} \), represents the points on the shadow manifold \( M_x \), where the coordinates of the \( k \)-th point are given by \( \underline{x}(k) = P[k, :] \).
To ensure consistency with the trajectory length \( \bar{L} \) of the shadow manifolds across different time series, we define the prediction target as $\mathbf{Y} = \hat{\mathbf X}[:,-\bar{L}:]$, where $\mathbf{Y} \in \mathbb{R}^{N \times \bar{L}}$. ${\mathbf{Y}}$ is reversed along the $\bar{L}$ dimension, the $n{\text{-th}}$ target time series is \( y(k) = \mathbf{Y}[n, :] \).

\paragraph{\ding{173} Estimate $y(k)$ within Multi-Manifold Space}
We use the points in the shadow manifold $M_x$ of one web service time series to predict the values $y(k)$ of another web service time series, denoted by $\hat{y}(k) \mid M_x$.
This process can validate the existence of causality between the $n{\text{-th}}$ and $m{\text{-th}}$ time series of $\hat{\mathbf{X}}$. 

We begin by locating the contemporaneous point $\underline{x}(k)$ in $M_x$, and find its $D+1$ nearest neighbors, denoting their time indices (from closest to farthest) by $t_{i_1},...,t_{i_{D+1}}$. 
Note that $D+1$ is the minimum number of points needed for a bounding simplex in an $D$-dimensional space. 
These neighbors are used to identify points in time series $y$ to estimate $y(k)$ from a locally weighted mean of the $y(t_{i_s})$ values:
\begin{equation}
    \hat{y}(k) \mid M_x = \sum_{s=1}^{D+1} w_{i_s} y(t_{i_s}),\nonumber
\end{equation}
where \( w_{i_s} \) represents the weight based on the distance between \( \underline{x}(k) \) and its \( s{\text{-th}} \) nearest neighbor. 
The weights \( w_{i_s} \) are determined by:
\begin{align}
    w_{i_s} &= \frac{u_{i_s}}{\sum_{j=1}^{D+1} u_{i_j}},\nonumber\\
    u_{i_s} &= \exp \left\{ -\frac{d[\underline{x}(k), \underline{x}(t_{i_s})]}{d[\underline{x}(k), \underline{x}(t_{i_1})]} \right\},\nonumber
\end{align}
where \( d[\underline{x}(k), \underline{x}(t_{i_s})] \) denotes the \textbf{Euclidean distance} between the two points in the shadow manifold $M_x$.

\paragraph{\ding{174} Momentum-Updated Correlation Coefficient Matrix}
The predictions are then compared with the ground truth values of the target web service time series to compute the correlation coefficient for each pair. 
As a result, we can obtain the causal correlation matrix: $ \mathbf{M} \in \mathbb{R}^{N\times N}$, where each element represents the causal relationship between a pair of web services.

To illustrate, we use a specific element of \( \mathbf{M} \) indexed as \( (m, n) \) to demonstrate the process for quantifying causal relationships. 
\( \mathbf{M}[m, n] \) quantifies the causal effect of the $n{\text{-th}}$ time series of $\hat{\mathbf{X}}$ on the $m{\text{-th}}$ time series of $\hat{\mathbf{X}}$
% service tim e series \( \mathbf{X}[n] \) on \( \mathbf{X}[m] \),
and can be calculated as follows:
\begingroup
\footnotesize
\begin{equation}
    \mathbf{M}[m, n] = \frac{\sum_{k=0}^{\bar{L}-1} \left( y(k) - \overline{y}(k) \right) \left( \hat{y}(k) - \overline{\hat{y}}(k) \right)}
    {\sqrt{\sum_{k=0}^{\bar{L}-1} \left( y(k) - \overline{y}(k) \right)^2 \sum_{k=0}^{\bar{L}-1} \left( \hat{y}(k) - \overline{\hat{y}}(k) \right)^2}}.\nonumber
\end{equation}
\endgroup

% To ensure consistent and stable representations, the $\mathbf{M}$ is then momentum update with previous iteration $\mathbf{M}^{iter-1} \in \mathbb{R}^{N \times N}$, and then softmax. 

% $\mathbf{M} = m * (Repeat into B for first dimension 
%   \mathbf{M}^{iter-1}) + (1- m) * \mathbf{M}$

% \widehat{\mathbf{M}}  \gets \text{Softmax}(\mathbf{M}, \text{dim}=-1)$,

% then the first batch size is averaged to have the general feature for $\mathbf{M}_{cc}$ for the $i$-th shadow manifold of each web service time series.

% ******

To ensure consistent and stable representations, the $\mathbf{M}$ matrix is updated using a momentum-based approach~\cite{he2020momentum} with the causal correlation matrix $\widetilde{\mathbf{M}}^{z-1} \in \mathbb{R}^{N \times N}$ from the previous iteration $z$-1, followed by a softmax operation for normalization:
\begin{align}
\mathbf{M}^{z} &= (1 - \lambda) \cdot \mathbf{M} + \lambda \cdot \widetilde{\mathbf{M}}^{z-1},\label{eq:causal_matrix_update}\\
\widehat{\mathbf{M}}^{z}(i) &= \text{Softmax}(\mathbf{M}^{z}),\nonumber
\end{align}
where \( z \) is the current iteration number, and $\lambda$ is the momentum value in the range $(0,1)$.
$\widehat{\mathbf{M}}^{z}(i) \in \mathbb{R}^{N \times N}$ captures causal relationships among the \( N \) web services with the  \( i \)-th shadow manifold of each web service time series.
% Also, $CCMPlus_representation$ incorporating the causal relationship among web services for the \( i \)-th shadow manifold is computed as follows:

% \( \boldsymbol{conv\_out\_ccm} \) with the shape \( (B, N, \bar{L}, D) \), is first reshape to change the last two dimensions, then use Linearlayer to map the dimension $\bar{L}$ into 1 , and squeeze it. 
% have intermediate_value

% After that, 

% (\text{Softmax}(\mathbf{M}, \text{dim}=-1) with shape (B, N, N) is multiply with above intermediate_value, after that, the CCMPlus_representation is generated.
\paragraph{\ding{175} Causality Enhanced Time Series Representation}
To incorporate causality among web services into the feature representation $\mathbf{h}_{ccm}(i)$ of time series, we use the causality score to weight the feature representation $\widehat{\mathbf{X}_{ccm}}$ of the shadow manifold embedding. Specifically, we first transpose and mapping $\mathbf{X}_{ccm}$ into $\widehat{\mathbf{X}_{ccm}} \in \mathbb{R}^{ N \times D} $, and then computed the causality-weighted representation:

\begin{align}
    \mathbf{h}_{ccm}(i) &= \widehat{\mathbf{M}}^{z}(i) \cdot \widehat{\mathbf{X}_{ccm}}, \label{eq:calculate_ccmplus_repre}\\
    \widehat{\mathbf{X}_{ccm}} &= \text{Transpose}(\mathbf{X}_{ccm})\cdot \mathbf{w}+b, \label{eq:calculate_conv_val}
\end{align}
where $\mathbf{h}_{ccm}(i) \in \mathbb{R}^{ N \times D}$ is the feature representation derived from the \( i \)-th shadow manifold. 
The CCMPlus representation $\mathbf{\widetilde{h}_{ccm}}$ and causal correlation matrix $\widetilde{\mathbf{M}}^{z}$ are computed by averaging the representations and matrices from all shadow manifolds:
\begin{align}
\mathbf{\widetilde{h}_{ccm}} &= \text{Mean}(\sum_{i=1}^{n }\mathbf{h}_{ccm}(i)), \label{eq:calcualte_ccmplus_re_iter}\\
\widetilde{\mathbf{M}}^{z} &= \text{Mean}(\sum_{i=1}^{n}\widehat{\mathbf{M}}^{z}(i)),\nonumber
\end{align}
where $\widetilde{\mathbf{M}}^{z} \in \mathbb{R}^{N \times N}$ and $\mathbf{\widetilde{h}_{ccm}} \in \mathbb{R}^{ N\times D}$ are the causal correlation matrix of the $z$-th iteration and the feature representation of the CCMPlus module respectively.

\subsection{Backbone Time Series Model}
\label{sec:backbone_ts_model}
The Backbone Time Series Model, denoted as \( \text{BTSM} \), processes the input time series embedding \( \mathbf{X} \in \mathbb{R}^{ N \times L \times C} \) and outputs a feature representation \( \mathbf{\widetilde{h}_{ts}} \in \mathbb{R}^{ N \times Q} \): $\mathbf{\widetilde{h}_{ts}} = \text{BTSM}(\mathbf{X})$,
which is concatenated with  \( \mathbf{\widetilde{h}_{ccm}} \) in the training or testing mode, to jointly predict the web service traffic time series.

% Based on the baselines evaluation performances on three real world web service traffic datasets across multiple prediction Granularity in Tables~\ref{tab:30T_performance}, \ref{tab:15T_performance}, \ref{tab:5T_performance}, and \ref{tab:1T_performance}, also considering the research community analysis~\cite{qiu2024tfb,wang2024deep}, the Timesnet and iTransformer are 2 good performing baselines among many SOTA time series models.

% So we use TimesNet and iTransformer as two Backbone Time Series Models to combine with CCMPlus (CCM+) module and verify brooadly the effectiveness of CCMPlus module. The resulting models are called CCM+iTransformer and CCM+TimesNet.

Based on the baseline evaluation performances across three real-world web service traffic datasets at multiple prediction granularities (Tables~\ref{tab:30T_performance}, 
\ref{tab:5T_performance}
  and 
  Tables~\ref{tab:15T_performance}, \ref{tab:1T_performance} 
  in Appendix) and insights from recent research~\cite{qiu2024tfb,wang2024deep}, TimesNet~\cite{wutimesnet} and iTransformer~\cite{liuitransformer} emerge as two strong-performing baselines among state-of-the-art time series models.
TimesNet identifies and utilizes multi-periodicity, decomposing temporal variations into intraperiod and interperiod components. The core module, TimesBlock, adaptively discovers periodicities and extracts features using a parameter-efficient inception block. iTransformer repurposes the Transformer architecture for time series forecasting by applying attention and feed-forward networks on inverted dimensions. It embeds time points as variate tokens, allowing attention to capture multivariate correlations and the feed-forward network to learn nonlinear variate-specific representations.

To broadly verify the effectiveness of the CCMPlus (CCM+) module, we integrate it with these two Backbone Time Series Models, resulting in two variant models, \textit{e.g.,} \textbf{CCM+iTransformer} and \textbf{CCM+TimesNet}.

\subsection{Optimization}
\label{sec:optimize}
For convenient combination with the Backbone Time Series Model and to improve generalization, the CCMPlus representation \( \mathbf{\widetilde{h}_{ccm}} \) is concatenated with the time series model feature representation \( \mathbf{\widetilde{h}_{ts}} \). 
In general, the whole procedure can be formalized as: $\hat{x} = \text{MLP}\left(\mathbf{\widetilde{h}_{ccm}} \mid \mathbf{\widetilde{h}_{ts}}\right)$,
where $\hat{x}$ is the predicted time series value, and MLP denotes the multilayer perception that projects the concatenated hidden feature to the final prediction.
Following previous work~\cite{liuitransformer,wutimesnet}, we employ the mean square error as the loss function to optimize the model parameters.
The overall training and inference algorithm can be found in Appendix Section~\ref{sec:algo}.

\begin{table*}[t]
\centering
\caption{Prediction performances averaged over three runs with prediction granularity $\alpha$ set as 30 minutes. The best result is marked in bold. The $t$-test conducted on both metrics indicates that the improvement is statistical significant (p-value < 0.001).}
\label{tab:30T_performance}
\resizebox{\textwidth}{!}{%
\begin{tabular}{lcccccccc}
\hline
\multicolumn{1}{l|}{30 Minutes} &
  \multicolumn{2}{c|}{Alibaba Group Traffic} &
  \multicolumn{2}{c|}{Microsoft Azure Traffic} &
  \multicolumn{2}{c|}{Ant Group Traffic} &
  \multicolumn{2}{c}{Overall Mean} \\ \hline
Method                                                      & MSE    & MAE     & MSE     & MAE    & MSE    & MAE    & MSE     & MAE    \\ \hline
MagicScaler~\cite{pan2023magicscaler} & 3.4909 & 0.5362  & 42.0655 & 0.8441 & 1.5513 & 1.0743 & 15.7026 & 0.8182 \\
OptScaler~\cite{zou2024optscaler}     & 3.5708 & 0.6147 & 33.7485 & 0.9459 & 1.3017 & 0.9421 & 12.8737 & 0.8342 \\
Llama3~\cite{touvron2023llama}        & 7.0570 & 1.1545  & 43.8206 & 1.8958 & 3.4346 & 1.5931 & 18.1041 & 1.5478 \\
TimeLLM~\cite{jintime}                & 3.4985 & 0.5339  & 17.2852 & 0.7088 & 1.5049 & 1.0560 & 7.4295  & 0.7662 \\
TimeMixer~\cite{wangtimemixer}        & 3.1429 & 0.5455  & 15.1801 & {0.6759} & 1.4036 & 0.9950 & 6.5755  & 0.7388 \\
iTransformer~\cite{liuitransformer}   & 3.1247 & 0.5428  & 19.5574 & 0.7933 & 1.4116 & 1.0010 & 8.0312  & 0.7790 \\
\textbf{CCM+iTransformer (ours)} &
  {3.0773} &
  \textbf{0.5098 ↓6.08\%} &
  \textbf{14.3191 ↓26.8\%} &
  \textbf{0.6482 ↓18.3\%} &
  {1.3162} &
  {0.9402} &
  \textbf{6.2375 ↓22.33\%} &
  \textbf{0.6994 ↓10.22\%} \\
TimesNet~\cite{wutimesnet}            & 3.1843 & 0.5406  & 16.6185 & 0.7177 & 1.4096 & 0.9989 & 7.0708  & 0.7524 \\
\textbf{CCM+TimesNet (ours)} &
  \textbf{3.0206 ↓5.14\%} &
  {0.5200} &
  {14.9237} &
  0.6791 &
  \textbf{1.2897 ↓8.51\%} &
  \textbf{0.9350 ↓6.40\%} &
  {6.4113} &
  {0.7114} \\ \hline
\end{tabular}%
}
\end{table*}

% \usepackage{graphicx}

% \usepackage{graphicx}
\begin{table*}[t]
\centering
\caption{Prediction performances averaged over three runs with prediction granularity $\alpha$ set as 5 minutes. The best result is marked in bold. The $t$-test conducted on both metrics indicates that the improvement is statistical significant (p-value < 0.001).}
\label{tab:5T_performance}
\resizebox{\textwidth}{!}{%
\begin{tabular}{lccccccll}
\hline
\multicolumn{1}{l|}{5 Minutes} &
  \multicolumn{2}{c|}{Alibaba Group Traffic} &
  \multicolumn{2}{c|}{Microsoft Azure Traffic} &
  \multicolumn{2}{c|}{Ant Group Traffic} &
  \multicolumn{2}{c}{Overall Mean} \\ \hline
Method                                                      & MSE    & MAE    & MSE     & MAE    & MSE    & MAE    & MSE    & MAE    \\ \hline
MagicScaler~\cite{pan2023magicscaler} & 3.2469 & 0.5176 & 9.3378  & 0.5473 & 1.5326 & 1.0640 & 4.7058 & 0.7096 \\
OptScaler~\cite{zou2024optscaler}     & 3.1070 & 0.4862 & 8.5807  & 0.5410 & 1.3214 & 0.9373 & 4.3364 & 0.6548 \\
Llama3~\cite{touvron2023llama}        & 7.3436 & 1.1474 & 11.5244 & 1.3587 & 3.3047 & 1.5606 & 7.3909 & 1.3556 \\
TimeLLM~\cite{jintime}                & 3.2588 & 0.5071 & 6.5393  & 0.5063 & 1.4974 & 1.0459 & 3.7652 & 0.6864 \\
TimeMixer~\cite{wangtimemixer}        & 2.7051 & 0.4818 & 3.0460  & 0.3890 & 1.3923 & 0.9819 & 2.3811 & 0.6176 \\
iTransformer~\cite{liuitransformer}   & 2.6458 & 0.4664 & 3.0367  & 0.3769 & 1.3939 & 0.9853 & 2.3588 & 0.6095 \\
CCM+iTransformer (ours)                                     & 2.5657 & 0.4460 & 2.9780  & 0.3662 & 1.3025 & 0.9398 & 2.2821 & 0.5840 \\
TimesNet~\cite{wutimesnet}            & 2.1681 & 0.3925 & 2.8696  & 0.3527 & 1.3923 & 0.9822 & 2.1433 & 0.5758 \\
\textbf{CCM+TimesNet (ours)} &
  \textbf{1.8103 ↓16.50\%} &
  \textbf{0.3394 ↓13.53\%} &
  \textbf{2.6347 ↓8.19\%} &
  \textbf{0.3150 ↓10.69\%} &
  \textbf{1.2871 ↓7.56\%} &
  \textbf{0.9287 ↓5.45\%} &
  \textbf{1.9107 ↓10.85\%} &
  \textbf{0.5277 ↓8.35\%} \\ \hline
\end{tabular}%
}
\end{table*}

\section{Experiments}
\subsection{Datasets}
We conduct experiments on three publicly available real-world web service traffic datasets from Ant Group\footnote{\url{https://huggingface.co/datasets/kashif/App_Flow/}}, Microsoft Azure\footnote{\url{https://github.com/Azure/AzurePublicDataset}}, and Alibaba Group\footnote{\url{https://github.com/alibaba/clusterdata}}. 
The Ant Group Traffic dataset includes 113 web services spanning a time range of 146 days. 
The Microsoft Azure Traffic dataset consists of 1,000 web services with a time range of 14 days. 
Similarly, the Alibaba Group Traffic dataset contains 1,000 web services, covering a total duration of 13 days. 
% The datasets utilized in this study are available at: \url{https://github.com/changtianluckyforever/CCMPlus}.

\subsection{Experiment Settings}
\subsubsection{Baselines}
\label{sec:baselines}
We evaluate our methods against the following baselines: 

\begin{itemize}
    \item \textbf{Large Language Models}: Llama3~\cite{touvron2023llama} and TimeLLM~\cite{jintime}.
    \item \textbf{Specialized Web Service Traffic Prediction Models}: MagicScaler~\cite{pan2023magicscaler} and OptScaler~\cite{zou2024optscaler}.
    \item \textbf{General Time Series Prediction Models}: TimesNet~\cite{wutimesnet}, TimeMixer~\cite{wangtimemixer}, and iTransformer~\cite{liuitransformer}.
\end{itemize}
More details about baselines can be found in the related work (\S\ref{sec:rlwk}).
These baselines represent the SOTA approaches in their respective categories. 
We apply the same experiment settings on these baselines and our methods to ensure fair comparison.

\subsubsection{Evaluation Metrics}  
We evaluate the model performances using Mean Squared Error (MSE) and Mean Absolute Error (MAE): 
\begin{equation}
\begin{array}{cc}
\text{MSE} = \frac{1}{n} \sum_{i=1}^n \left( y_i - \hat{y}_i \right)^2, & 
\text{MAE} = \frac{1}{n} \sum_{i=1}^n \left| y_i - \hat{y}_i \right|, \nonumber
\end{array}
\end{equation}
where \( y_i \) and \( \hat{y}_i \) denote the true and predicted values, respectively, and \( n \) is the number of samples in the test set.

\subsubsection{Implementation Details}
The Llama3 (8B) baseline is fine-tuned using LoRA (rank is set as 16). 
The $\tau_w$ is set as 100, and \( \tau \in [1, 2, 3, 4] \).
The momentum value $\lambda$ is 0.5. The channels of input and output of the convolution network are $C$ and $D$,
$C$ and $D$ are set as $16$ and $32$, respectively. 
The batch size \( B \) is 8, and the Adam optimizer is configured with a learning rate of 0.000001. 
The input length \( L \) is 168. 
Training is performed on an H100 GPU. 
The model is trained for 15 epochs with an early stopping patience of 5 epochs based on validation performance.
The prediction granularity parameter, $\alpha$, corresponds to the prediction horizon, distinguishing between long-term and short-term forecasts. $\alpha$ has following values: 1, 5, 15 and 30.
% \subsubsection{Research Questions}
% The following research questions are investigated: 1) What are the overall performances and prediction accuracies of the models across different granularities? (Section~\ref{sec:performance_analysis}) 2) How does the CCMPlus module impact prediction performance? (Section~\ref{sec:ablation_study})
% 3) How do the methods perform under varying hyperparameter settings? (Section~\ref{sec:hyperparameter_analysis})

\subsection{Prediction Performance Analysis}
\label{sec:performance_analysis}

\textbf{Performance Comparisons.} To evaluate the performance of CCMPlus (CCM+), we compare two variants of our proposed framework, CCM+TimesNet and CCM+iTransformer, against the baselines. 
As shown in Table~\ref{tab:30T_performance}, we report the prediction results on three datasets, with the prediction granularity $\alpha$~(explained in Eq.~\ref{eq:define}) set to 30 minutes.
Among all the baselines, Llama3 performs the worst. 
While large language models have demonstrated effectiveness in reasoning over sequential data, web traffic exhibits significantly higher volatility, posing substantial challenges for plain LLMs to accurately forecast future values. 
In contrast, TimeLLM inherently treats time series patches as tokens and employs task-specific prompt embeddings for forecasting. 
Although it captures patch correlations, it still underperforms on dynamic web traffic data compared to methods specifically tailored for time series analysis.
Magicscaler and OptScaler, in particular, are designed for service workload prediction. While these models account for the volatility inherent in traffic time series, they fail to explicitly capture the underlying seasonal and trend components, which are crucial for predicting web service traffic driven by human behavior. 

TimeMixer addresses this limitation by decomposing time series into seasonal and trend components across multiple periodicities, enabling it to learn decomposed temporal patterns ranging from fine-grained to macro-level perspectives. 
As a result, TimeMixer achieves superior prediction performance compared to both LLM-based methods and specialized workload prediction models.
TimesNet leverages Fourier transforms to derive more adaptive periodicity terms, further reducing prediction errors. 
On the other hand, iTransformer treats independent time series as tokens, learning both temporal and cross-dimensional correlations by modeling token sequences, and achieves performance comparable to TimesNet. 
However, all the above methods focus solely on internal temporal patterns while neglecting external causal relationships across multiple time series. 
To address this limitation, we integrate CCMPlus with TimesNet and iTransformer—the two best-performing baseline models—to create two new variants.
As observed, while TimesNet and iTransformer employ carefully designed architectures and achieve SOTA performance, integrating CCMPlus leads to further improvements. 
This highlights the importance of capturing the causality inherent in web service traffic to accurately predict traffic volume. 
Furthermore, CCMPlus effectively models this causality, contributing to enhanced predictive performance.

\begin{table*}[tbp]
\centering
\caption{Hyperparameter analysis of $C$, $D$, and $\tau$ on the Microsoft Azure Traffic dataset with 5‑minute prediction granularity using CCM+TimesNet. The results are averaged over three experiments.}
\label{tab:hyperparameter_analysis_1}
\resizebox{\textwidth}{!}{%
\begin{tabular}{ccccccccc}
\hline
$C$ & MSE    & MAE    & $D$ & MSE     & MAE     & $\tau$        & MSE     & MAE     \\ \hline
12                & 2.6873 & 0.3178 & 28                 & 2.8307  & 0.3149  & {[}1, 2, 3{]}       & 2.7945 & 0.3157 \\
14                & 2.8256 & 0.3385 & 30                 & 2.6894 & 0.3223 & {[}1, 2, 3, 4{]} (ours)    & 2.6347 & 0.3150 \\
16 (ours)                & 2.6347 & 0.3150 & 32 (ours)                & 2.6347 & 0.3150 & {[}1, 2, 3, 4, 5{]} & 2.6531 & 0.3144 \\
18                & 2.7078 & 0.3170 & 34                 & 2.6357  & 0.3143 &                     &         &         \\
20                & 2.6311 & 0.3117 & 36                 & 2.7461 & 0.3273 &                     &         &         \\ \hline
\end{tabular}%
}
\end{table*}

\textbf{Prediction Granularity Analysis.} Prediction granularity is an important factor for traffic forecasting. 
To assess this, we vary the prediction granularity $\alpha$ in the time series in \{1, 5, 15, 30\} minutes and compare the model performances in Table~\ref{tab:1T_performance}, \ref{tab:5T_performance}, \ref{tab:15T_performance} and \ref{tab:30T_performance}, respectively. (results for $\alpha=1$ and $\alpha=15$ are in Appendix~\S\ref{sec:add_exp}.)
Note that the more fine-grained time interval setting would generate more data samples.
First, we observe that CCMPlus consistently improves the SOTA models (TimesNet and iTransformer), underscoring the importance of considering causality among service traffic patterns. 
Notably, CCMPlus demonstrates flexibility in integrating with various SOTA backbone models, highlighting its potential for performance enhancement.
Second, the performance improvements diminish for the 1-minute prediction granularity setting. 
This is because service traffic is highly dynamic and volatile, and fine-grained time series data often lack a sufficient time duration to accumulate information necessary for capturing reliable causal relationships. Nevertheless, CCMPlus still achieves the best performance, further demonstrating its superiority in extracting hidden causal relationships and boosting prediction accuracy even under challenging conditions.
We conduct a $t$-test~\cite{bhattacharya2002median} under both metrics shows that the improvement of our method over iTransformer and TimesNet is significant ($p$-value < 0.001).

\subsection{Ablation Study}
\label{sec:ablation_study}

\begin{table}[tbp]
\centering
\caption{Ablation study on the Microsoft Azure Traffic dataset with time granularity $\alpha$ set as 5 minutes.}
\label{tab:ablation_study}
\resizebox{0.9\columnwidth}{!}{%
\begin{tabular}{lcc}
\hline
Method                         & MSE    & MAE    \\ \hline
w/o BTSM & 2.9981 & 0.3627 \\
w/o MME\&CER             & 2.8696 & 0.3527 \\
w/o CER             & 2.7329 & 0.3320 \\
\midrule
CCM+TimesNet (ours)             & 2.6347 & 0.3150 \\
\hline
\end{tabular}%
}
\end{table}

% CCMPlus model consistently outperforms multiple baselines across three datasets. 
The proposed model consists of the CCMPlus module~(\S\ref{sec:ccmplus_module}) and BTSM module~(\S\ref{sec:backbone_ts_model}), where the CCMPlus encompasses of two key internal representations, i.e., the multi-manifold embedding and causality enhanced time-series representation~(\ding{172} and \ding{175} introduced in \S\ref{sec:ccmplus_module}).
To assess the contribution of these modules and representations, we conduct a detailed ablation study using CCM+TimesNet on the most challenging dataset, Microsoft Azure Traffic. 
Table~\ref{tab:ablation_study} presents the impact of model variants built on these modules and representations, yielding the following insights: (\textbf{1}) Excluding the Backbone Time Series Model (w/o BTSM) reduces performance. The deep learning-based backbone extracts essential seasonal and trend-related temporal features, which are crucial for accurate web service traffic forecasting.  
(\textbf{2}) By comparing the results when both Multi-Manifold Embedding~(MME) and Causality Enhanced Representations~(CER) are removed with those when only CER is removed, we can observe that MME provides diverse time-sensitive representations from multiple shadow manifold spaces formed by different time lags and embedding dimension values, which can lead to performance improvement for the advanced BTSM model;
(\textbf{3}) Our complete model (i.e., BTSM+MME+CER) performs better than the variant without CER, which demonstrates that CER can generate feature representations that capture causal relationships among web services, enhancing traffic prediction accuracy.

\subsection{Hyperparameter Tuning}
Table~\ref{tab:hyperparameter_analysis_1} presents the hyperparameter analysis, evaluating the impact of the following key parameters:
(1) \( C \) defines the embedding dimension of the input time series \( \hat{\mathbf{X}} \). As shown in Table~\ref{tab:hyperparameter_analysis_1}, the model maintains stable performance across values, demonstrating robustness. Increasing this dimension enhances feature richness, but for efficiency, we set \( C = 16 \).
(2) \( D \) defines the embedding dimension of shadow manifold points. A small value may miss features, while a large one can introduce noise. To balance representation quality and efficiency, we set \( D = 32 \).
(3) The values in \( \tau \) define the dilation parameter in Conv1D, controlling the number of shadow manifolds. Increasing this number enhances feature extraction but incurs higher computational costs. To balance accuracy and efficiency, we set \( \tau = [1,2,3,4] \).

% As shown in Table~\ref{tab:hyperparameter_analysis_1}, momentum value \( m \) stabilizes the causal correlation matrix by integrating current and past iterations, as defined in Equation~\ref{eq:causal_matrix_update}. Setting $m$ loses general features, while \( m = 0.8 \) reduces adaptability. We choose \( m = 0.5 \) for a balance between stability and adaptability.

\begin{figure}[tbp]
  \centering
  \subfloat[The causal relationship heatmap of web service A.]{
\includegraphics[width=0.95\linewidth]{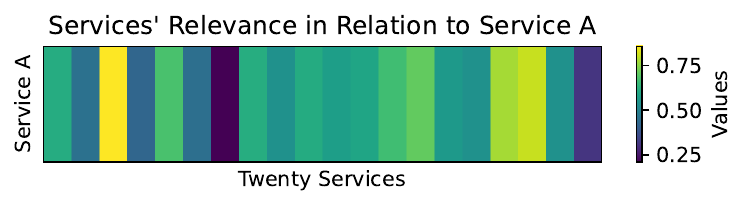}
  \label{fig:services_heatmap}}

\subfloat[Traffic time-series plots of web service A and service B.]{
\includegraphics[width=0.95\columnwidth]{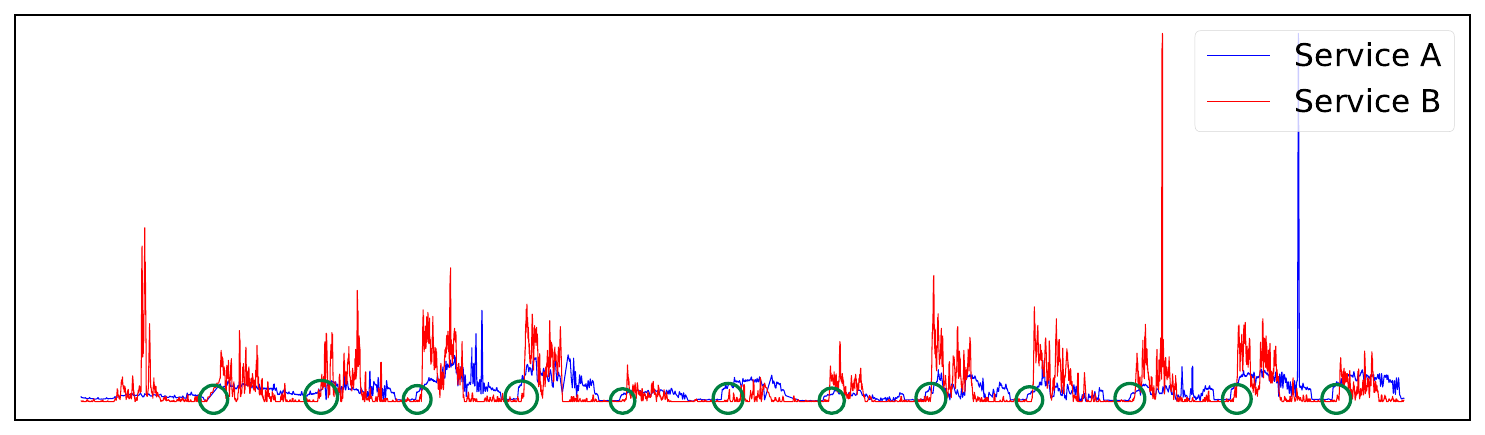}
  \label{fig:services_heatmap_AB}
}

\subfloat[Traffic time-series plots of web service A and service C.]{
\includegraphics[width=0.95\columnwidth]{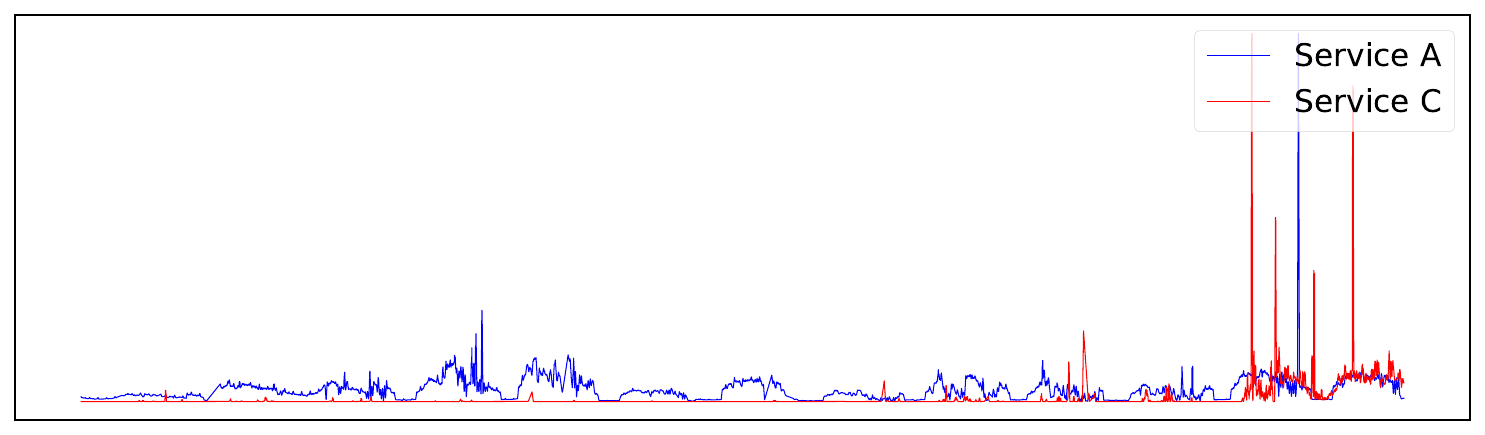}
  \label{fig:services_heatmap_AC}
}

\caption{The heatmap of web service A's causal relationships and the plots of the service A, B and C with the highest and lowest causal relationship, respectively.}
\vspace{0.5cm}
\label{fig:example}
\end{figure}

\subsection{Case Study}

\label{sec:hyperparameter_analysis}
% \begin{figure*}[tbh!]
%   \centering
% \includegraphics[width=\textwidth]{figures/services_heatmap_AB.pdf}
%   \caption{The causal relationship between Web service A and service B}
%   \label{fig:services_heatmap_AB}
% \end{figure*}
Using the Alibaba Group Traffic dataset, we compute causal relationships between web services with the CCMPlus module. 
To visualize, we compare Service A with twenty randomly sampled services and plot the causal correlation heatmap (Figure~\ref{fig:services_heatmap}), where Service A exhibits a strong causal correlation with the third service, referred to as Service B, and the weakest causal correlation with the seventh service, referred to as Service C. 
To further analyze this relationship, we plot the traffic time series of Services A, B and C over the same period (Figure~\ref{fig:services_heatmap_AB} and \ref{fig:services_heatmap_AC}). 
Figure~\ref{fig:services_heatmap_AB} indicate the causality correlation between service A and B, where a increase of A prompt the lagged increase of B~(as marked in green circle), which exhibits a high causality score.
From Figure~\ref{fig:services_heatmap_AC}, we can observe that A and C show no discernible correlation, and the traffic of service C is steady and independent with A, which is also aligned with the heatmap results.

% \subsection{Error Analysis *consider remove?}

% \begin{table}[tbh!]
% \centering
% \caption{Error analysis of the 10 services with the largest and smallest MSE values, evaluated in terms of ADF scores. The analysis was conducted on the Microsoft Azure Traffic dataset with a 5T time interval using the CCM+TimesNet.}
% \label{tab:error_analysis}
% \resizebox{0.7\linewidth}{!}{%
% \begin{tabular}{cc}
% \toprule
% \textbf{Top 10 Services Category} & \textbf{Averaged ADF Score} \\ \midrule
% Smallest MSE & -20.9963 \\
% Largest MSE & -6.8813 \\ \bottomrule
% \end{tabular}%
% }
% \end{table}

% We analyze errors on the Azure Traffic dataset (5T interval) using the Augmented Dickey-Fuller (ADF) test to assess stationarity. A lower (more negative) ADF score indicates higher stationarity. Examining the 10 web services with the highest and lowest MSE values, we find that larger MSE losses correspond to higher averaged ADF scores. This suggests that volatile time series are harder to model due to their complex temporal patterns, posing challenges for time series forecasting.

\section{Conclusion}
We propose the CCMPlus module, which captures causal relationships between web services to improve web service traffic prediction.
CCMPlus first constructs multi-manifold for each web service traffic time series. 
It then estimates target time series from the multi-manifold space of other web services. 
By comparing these estimations with the ground truth, a causal correlation matrix is computed and used to generate the CCMPlus representation, quantifying inter-service causal dependencies. 
% The causal matrix is then applied to generate the CCMPlus representation, which encodes causal relationships. 
This representation is concatenated with the time series model’s output, bridging the gap in causal inference and enhancing prediction performance.
We evaluate CCMPlus-integrated models on real-world web service traffic datasets, which demonstrates that CCMPlus consistently improves web service traffic prediction.
% \section{Ethics Statement}
% We introduce CCMPlus, a neural network module for extracting causal relationship features across services. This work has no direct negative ethical impacts.
%%%%%%%%%%%%%%%%%%%%%%%%%%%%%%%%%%%%%%%%%%%%%%%%%%%%%%%%%%%%%%%%%%%%%%%%

%%% Use this environment to include acknowledgements (optional).
%%% This will be omitted in doubleblind mode.

% \begin{ack}
% By using the \texttt{ack} environment to insert your (optional) 
% acknowledgements, you can ensure that the text is suppressed whenever 
% you use the \texttt{doubleblind} option. In the final version, 
% acknowledgements may be included on the extra page intended for references.
% \end{ack}

%%%%%%%%%%%%%%%%%%%%%%%%%%%%%%%%%%%%%%%%%%%%%%%%%%%%%%%%%%%%%%%%%%%%%%%%

%%% Use this command to include your bibliography file.

\bibliography{mybibfile}

\newpage
\section*{Appendix}
\begin{appendix}
We offer some technical details and reproducibility-related information as supplementary materials to help readers understand and reproduce our model.

\section{Cross Convergence Mapping}
\label{sec:ccm_procedure}
Convergent Cross Mapping (CCM) theory is a classical algorithm proposed to detect causal relationships between species. 
Inspired by this algorithm and preliminary experiments where web services show latent causality, we leverage CCM to enhance traffic prediction.
To help readers better understand the CCM, we provide more details in this section.

\begin{figure}[h]
  \centering
  \includegraphics [width=\columnwidth]{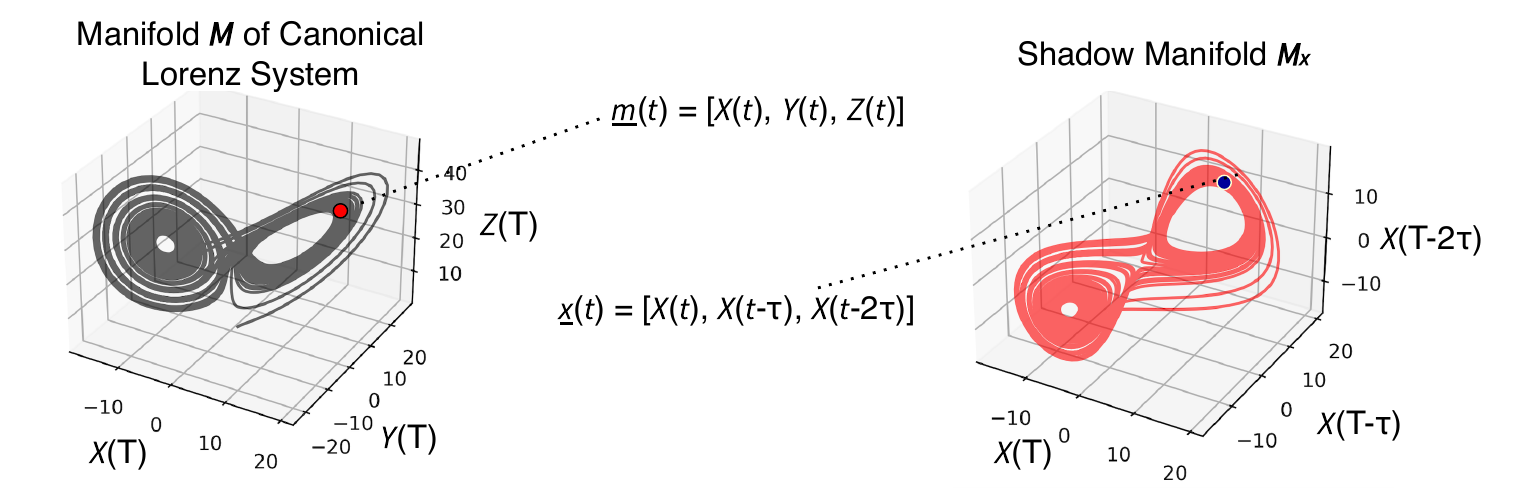}
  \caption{The CCM theory is explained using the canonical Lorenz system. This figure exemplifies the manifold \(M\) and its corresponding shadow manifold \(M_x\).}
\label{fig:ccm_lorenz_system}
  % \vspace{0.5cm}
\end{figure}

The exposition of CCM is often illustrated using the context of the Lorenz system.
As depicted in Figure~\ref{fig:ccm_lorenz_system}, the trajectory of the Lorenz system forms a manifold \( M \) in the state space. 
This manifold \( M \) consists of a collection of points that represent all possible states of the Lorenz system over time, with these points connected to create a structured geometric space. 
The manifold, also referred to as the attractor, encompasses all trajectories and potential states \( \underline{m}(t) \) of the system. Each state \( \underline{m}(t) \) corresponds to a point in \( M \), represented by the coordinate vector \( \underline{m}(t) = [X(t), Y(t), Z(t)] \).

\begin{figure}[h]
  \centering
  \includegraphics[width=0.99\columnwidth]{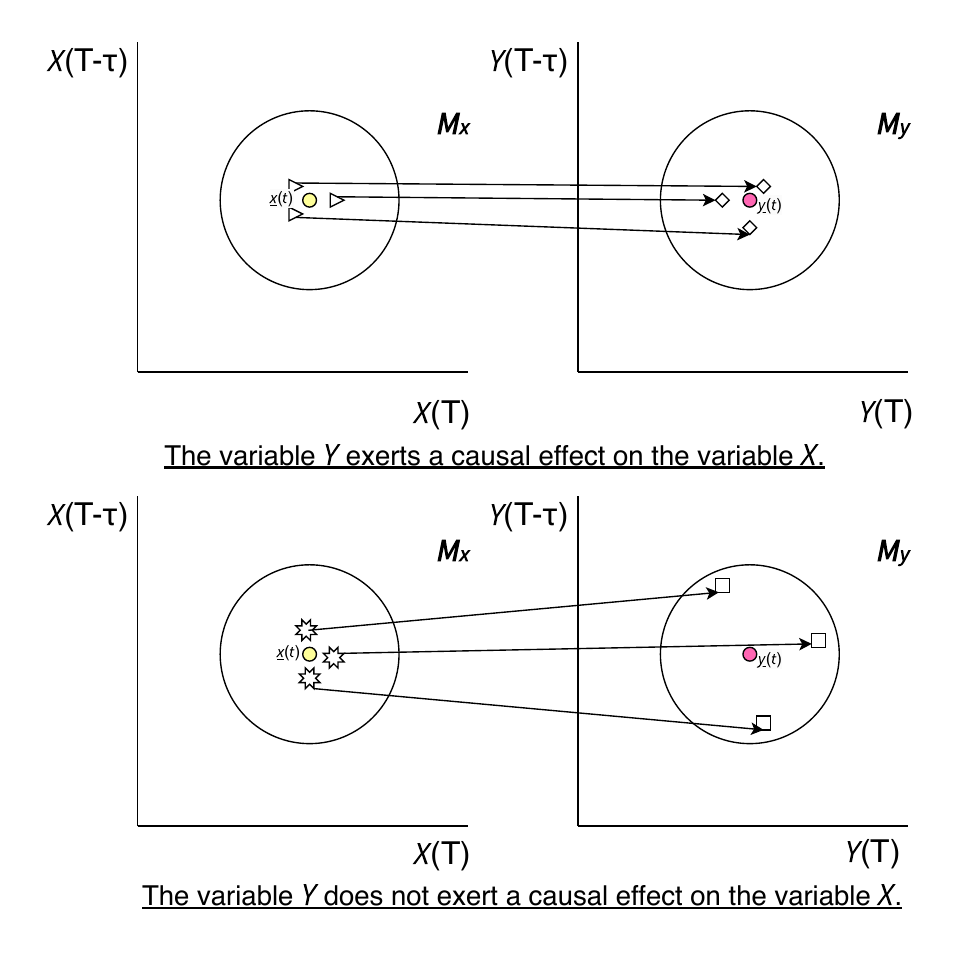} 
  \caption{The point \( \underline{y}(t) \) in the manifold \( M_y \) corresponds to the contemporaneous point in time \( \underline{x}(t) \) in the manifold \( M_x \).}
  \label{fig:ccm_cross_mapping}
  \vspace{0.5cm}
\end{figure}

\begin{figure}[h]
  \centering
  \includegraphics[width=0.8\columnwidth]{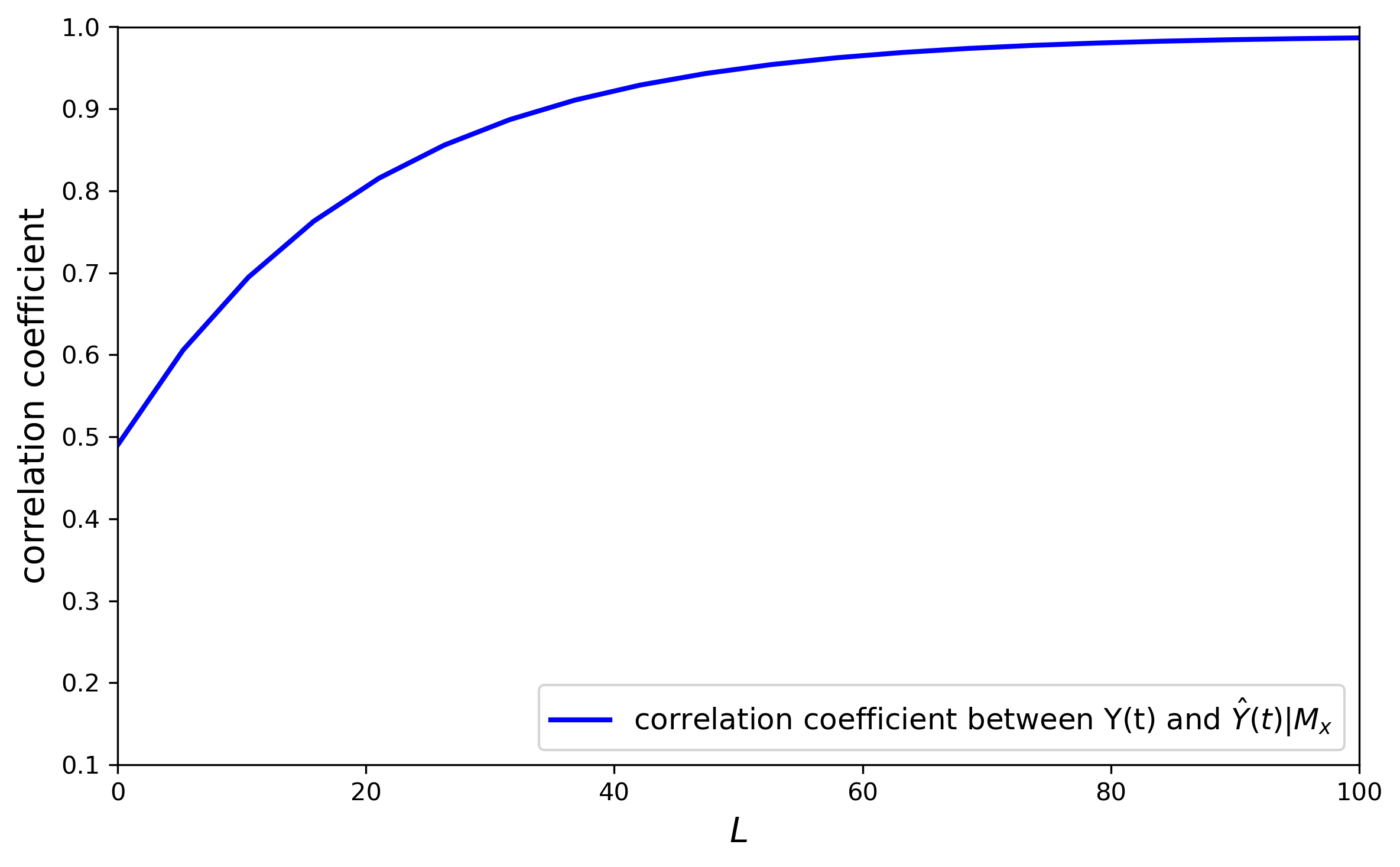} 
  \caption{Convergent predictability as the time series length increases, assuming $Y$ has a causal effect on $X$.
}
  \label{fig:ccm_convergence}
  \vspace{0.5cm}
\end{figure}

The CCM procedure for detecting whether variable \( Y \) has causal effects on \( X \) consists of four key steps:

\begin{itemize}
    \item \textbf{Step 1: Construct the shadow manifold \( M_x \).}  
    Consider two time-evolving variables \( X(k) \) and \( Y(k) \) of length \( L \), where the time index \( k \) ranges from \( 1 \) to \( L \). The shadow manifold \( M_x \) is constructed by forming lagged coordinate vectors:
    \[
    \underline{x}(t) = [X(t), X(t-\tau), X(t-2\tau), \dots, X(t-(E-1)\tau)],
    \]
    for \( t = 1+(E-1)\tau \) to \( t=L \). Here, \( \tau \) represents the time lag, and \( E \) denotes the embedding dimension.

    \item \textbf{Step 2: Identify nearest neighbors in \( M_x \).}  
    To estimate \( Y(t) \) for a specific \( t \) in the range \( 1+(E-1)\tau \) to \( L \), use the shadow manifold \( M_x \). Denote the estimated value as \( \hat{Y}(t) \mid M_x \). Begin by locating the contemporaneous lagged coordinate vector \( \underline{x}(t) \) in \( M_x \), and find its \( E+1 \) nearest neighbors. Note that $E+1$ is the minimum number of points needed for a bounding simplex in an $E$-dimensional space. Let the time indices of these neighbors (ranked by proximity) be \( t_1, t_2, \dots, t_{E+1} \). The nearest neighbors of \( \underline{x}(t) \) in \( M_x \) are therefore denoted by \( \underline{x}(t_i) \), where \( i = 1, \dots, E+1 \).

    \item \textbf{Step 3: Estimate \( Y(t) \) using locally weighted means.}  
    The time indices \( t_1, t_2, \dots, t_{E+1} \) corresponding to the nearest neighbors of \( \underline{x}(t) \) are used to identify points in the variable \( Y(k) \). These points are then used to estimate \( Y(t) \) through a locally weighted mean of the \( E+1 \) values \( Y(t_i) \):
    \begin{equation}
    \hat{Y}(t) \mid M_X = \sum_{i=1}^{E+1} w_i Y(t_i),
    \nonumber
    \end{equation}
    where \( w_i \) represents the weight based on the distance between \( \underline{x}(t) \) and its \( i{\text{-th}} \) nearest neighbor in \( M_x \), and \( Y(t_i) \) are the contemporaneous values of variable \( Y(k) \). The weights \( w_i \) are determined by:
    \begin{equation}
    w_i = \frac{u_i}{\sum_{j=1}^{E+1} u_j},
    \nonumber
    \end{equation}
where
    \begin{equation}
    u_i = \exp \left\{ -\frac{d[\underline{x}(t), \underline{x}(t_i)]}{d[\underline{x}(t), \underline{x}(t_1)]} \right\}.
    \nonumber
    \end{equation}
    Here, \( d[\underline{x}(t), \underline{x}(t_i)] \) denotes the Euclidean distance between the two vectors.

    \item \textbf{Step 4: Calculate the correlation coefficient \( r \).}  
    Finally, calculate the correlation coefficient \( r \) between \( \hat{Y}(t) \) and \( Y(t) \), where \( t \) ranges from \( 1+(E-1)\tau \) to \( L \):
    \begingroup
\footnotesize
\begin{equation}
    r = \frac{\sum_{t=1+(E-1)\tau}^L \left( Y(t) - \overline{Y}(t) \right) \left( \hat{Y}(t) - \overline{\hat{Y}}(t) \right)}
    {\sqrt{\sum_{t=1+(E-1)\tau}^L \left( Y(t) - \overline{Y}(t) \right)^2 \sum_{t=1+(E-1)\tau}^L \left( \hat{Y}(t) - \overline{\hat{Y}}(t) \right)^2}}.
\nonumber
\end{equation}
\endgroup
    If variable \( Y \) has causal effects on \( X \), \( \hat{Y}(t) \) will converge to \( Y(t) \) as the observation period increases. In ideal cases, the correlation coefficient \( r \) will approach 1.
\end{itemize}

\section{Overall Training Algorithm}
\label{sec:algo}
Algorithm~\ref{alg:ccmplus_module} presents the training algorithm for our CCMPlus module corresponding to the procedure introduced in \S\ref{sec:ccmplus_module}, which produce the causality enhanced representation $\mathbf{\widetilde{h}_{ccm}}$ and causal correlation matrix $\widetilde{\mathbf{M}}^{z}$, respectively.

\begin{algorithm*}[h]

\caption{Algorithmic Procedure for the CCMPlus Module}
\label{alg:ccmplus_module}
\begin{algorithmic}[1]

\Require $\tau = [\tau_1, \ldots, \tau_n]$ : list of time lagged parameters, $\tau_w$ : time window length, 
$L$: input time series length,
$N$: the number of web service within the input,
$C$: the number of input channels for Conv1D,
$D$: the number of output channels of the Conv1D,
$\hat{\mathbf{X}} \in \mathbb{R}^{ N\times L}$: input time series,
$\mathbf{X}\in \mathbb{R}^{ N \times L \times C}$: input time series embedding,
$\mathbf{M} \in \mathbb{R}^{N\times N}$: causal correlation matrix,
$m$: the momentum value used for updating the causal correlation matrix, constrained within the range \( (0, 1) \), $\text{training\_flag} \in \{\text{True}, \text{False}\}$ : training or testing mode.
% \Ensure \textbf{Output} $h_{ccm}^{iter}$ and $\mathbf{M}_{cc}^{iter}$
\Ensure \textbf{Output} $\mathbf{\widetilde{h}_{ccm}}$ and $\widetilde{\mathbf{M}}^{z}$
\State Initialize $\mathbf{E}$ as an empty list.
\For{each $\tau_i$ in $\tau$}
    \State $E_i \gets \left\lfloor \dfrac{\tau_w}{\tau_i} \right\rfloor$
    \If{$E_i$ is even}
        \State $E_i \gets E_i - 1$
    \EndIf
    \State Append $E_i$ to $\mathbf{E}$
\EndFor

\For{$z \gets 0, \ldots, Z$}
    % \State $\text{multi\_space\_representations} = []$, $\text{multi\_space\_corrs} = []$ 
    \State $\mathcal{H} = []$, $\mathcal{C} = []$ 
    \For{$i \gets 1 \ldots n$}
        \State $\tau_i \gets \tau[i-1]$, \quad $E_i \gets \mathbf{E}[i-1]$
        \If{$(L - \tau_i \times (E_i - 1)) < 0$}
            \State \textbf{continue} 
        \EndIf

        \State Reverse the order of $\mathbf{X}$ along the $L$ dimension. Then reshape $\mathbf{X}$ to $\bar{\mathbf{X}} \in \mathbb{R}^{ N\times C \times L}$
 
        \State $\mathbf{X}_{conv} = \text{Conv1D}\bigl(\bar{\mathbf{X}};\,\text{kernel\_size}=E_i,\text{dilation}=\tau_i,
% \text{in\_channels}=C_{in},\text{out\_channels}=C_{out} 
        \bigr)$, 
        $\mathbf{X}_{conv} \in \mathbb{R}^{ N \times D \times  \bar{L} }$ 
        % $L_{out} = L_x - \tau_i \cdot (E_i-1))$ 

        \If{$\text{training\_flag} = \text{True}$}
            \State Reshape $\mathbf{X}_{conv}$ to $\mathbf{X}_{ccm}\in \mathbb{R}^{N \times \bar{L} \times D}$, $\mathbf{Y} \gets \hat{\mathbf{X}}[:,-\bar{L}:]$, ${\mathbf{Y}}$ is reversed along the $\bar{L}$ dimension

            \State Compute the distance matrix $\mathcal{D} \gets \text{PairwiseDistances} (\mathbf{X}_{ccm} \mathbf{X}_{ccm})$
            \State Find the nearest neighbor's indices of each web service $\mathcal{I} \gets \text{argsort}(\mathcal{D}, \text{dim}=-1)[:, :, 1:(D+2)]$

            \State $d \gets \text{Gather}(\mathcal{D}, \mathcal{I})$

            \State $u_z \gets \exp\!\Bigl(-\,\dfrac{d}{d[ :, :, 0{:}1] + \epsilon}\Bigr)$, 
        $w_z \gets \dfrac{u_z}{\sum(u_z) + \epsilon}$

            \State $Y_{nearest} \gets \text{Gather}( \mathbf{Y}, \mathcal{I} )$, $\hat{Y} \gets \sum (w_z \times Y_{nearest})$

            \State $corr \gets \dfrac{\mathrm{Cov}(\hat{Y},\, \mathbf{Y})}{\sigma(\hat{Y})\,\sigma(\mathbf{Y}) + \epsilon}$,  $\mathbf{M} \gets corr^\top$ 
            \If{$z$ >0}
                \State $\widehat{\mathbf{M}}^{z}(i) \gets \lambda * \widetilde{\mathbf{M}}^{z-1} + (1- \lambda) * \mathbf{M}$
            \Else 
            \State $\widehat{\mathbf{M}}^{z}(i) \gets  \mathbf{M}$
            
            \EndIf

        \Else
            \State $\widehat{\mathbf{M}}^{z}(i) \gets \widetilde{\mathbf{M}}^{z-1}$
            
        \EndIf

        \State  $\widehat{\mathbf{X}_{ccm}} \gets \text{LinearProjection}(\text{Transpose}(\mathbf{X}_{ccm}))$
        \State
        $\mathbf{h}_{ccm}(i) = \widehat{\mathbf{M}}^{z}(i) \cdot \widehat{\mathbf{X}_{ccm}}$

        \State Append $h_{ccm}(i)$ to $\mathcal{H}$; Append $\widehat{\mathbf{M}}^{z}(i)$ to $\mathcal{C}$
    \EndFor

    \State $\mathbf{\widetilde{h}_{ccm}} \gets \text{Mean}(\text{Stack}(\mathcal{H}), \text{dim}=0)$
    
    \State $\widetilde{\mathbf{M}}^{z} \gets \text{Mean}(\text{Stack}(\mathcal{C}), \text{dim}=0)$
    
\EndFor
\end{algorithmic}
\end{algorithm*}

\section{Additional Performance Comparison Results}
\label{sec:add_exp}
In this part, we provide additional results of the prediction granularity experiments~($\alpha=1$ and $\alpha=15$).
It show that with our proposed CCMPlus module, the SOTA models can achieve further improvements on various prediction granularity setting across all the datasets.
The $t$-test also demonstrates the improvements are statistical significant.

\begin{table*}[t]
\centering
\caption{Prediction performances averaged over three runs with prediction granularity $\alpha$ set as 15 minutes. The best result is marked in bold. The $t$-test conducted on both metrics indicates that the improvement is statistical significant (p-value < 0.001).}
\label{tab:15T_performance}
\resizebox{\textwidth}{!}{%
\begin{tabular}{lcccccccc}
\hline
\multicolumn{1}{l|}{15 Minutes} &
  \multicolumn{2}{c|}{Alibaba Group Traffic} &
  \multicolumn{2}{c|}{Microsoft Azure Traffic} &
  \multicolumn{2}{c|}{Ant Group Traffic} &
  \multicolumn{2}{c}{Overall Mean} \\ \hline
Method                                                      & MSE     & MAE    & MSE     & MAE    & MSE    & MAE    & MSE    & MAE    \\ \hline
MagicScaler~\cite{pan2023magicscaler} & 3.3695  & 0.5262 & 19.2405 & 0.6764 & 1.5044 & 1.0544 & 8.0381 & 0.7523 \\
OptScaler~\cite{zou2024optscaler}     & 3.4188 & 0.5950 & 17.0180 & 0.7560 & 1.3069 & 0.9503 & 7.2479 & 0.7671 \\
Llama3~\cite{touvron2023llama}        & 7.4170  & 1.1452 & 12.6471 & 1.5288 & 3.3408 & 1.5700 & 7.8016 & 1.4147 \\
TimeLLM~\cite{jintime}                & 3.3954  & 0.5242 & 6.7723  & 0.6217 & 1.5176 & 1.0579 & 3.8951 & 0.7346 \\
TimeMixer~\cite{wangtimemixer}        & 2.8915  & 0.5004 & 5.8507  & 0.5522 & 1.3962 & 0.9900 & 3.3795 & 0.6809 \\
iTransformer~\cite{liuitransformer}   & 2.8854  & 0.5118 & 5.7501  & 0.5310 & 1.4017 & 0.9938 & 3.3457 & 0.6789 \\
CCM+iTransformer (ours)                                     & 2.8374  & 0.4932 & 5.0156  & 0.4582 & 1.3127 & 0.9368 & 3.0552 & 0.6294 \\
TimesNet~\cite{wutimesnet}            & 2.8635  & 0.4904 & 5.0550  & 0.4641 & 1.3962 & 0.9876 & 3.1049 & 0.6474 \\
\textbf{CCM+TimesNet (ours)} &
  \textbf{2.7151 ↓5.18\%} &
  \textbf{0.4823 ↓1.65\%} &
  \textbf{4.7434 ↓6.16\%} &
  \textbf{0.4422 ↓4.72\%} &
  \textbf{1.2951 ↓7.24\%} &
  \textbf{0.9199 ↓6.85\%} &
  \textbf{2.9179 ↓6.02\%} &
  \textbf{0.6148 ↓5.04\%} \\ \hline
\end{tabular}%
}
\end{table*}

\begin{table*}[t]
\centering
\caption{Prediction performances averaged over three runs with prediction granularity $\alpha$ set as 1 minute. The best result is marked in bold. The $t$-test conducted on both metrics indicates that the improvement is statistical significant (p-value < 0.001).}
\label{tab:1T_performance}
\resizebox{\textwidth}{!}{%
\begin{tabular}{lcccccccc}
\hline
\multicolumn{1}{l|}{1 Minute} &
  \multicolumn{2}{c|}{Alibaba Group Traffic} &
  \multicolumn{2}{c|}{Microsoft Azure Traffic} &
  \multicolumn{2}{c|}{Ant Group Traffic} &
  \multicolumn{2}{c}{Overall Mean} \\ \hline
Method                                                      & MSE    & MAE             & MSE             & MAE    & MSE    & MAE    & MSE    & MAE    \\ \hline
MagicScaler~\cite{pan2023magicscaler} & 2.7799 & 0.4783          & 6.7159          & 0.4748 & 1.9403 & 1.1674 & 3.8120 & 0.7068 \\
OptScaler~\cite{zou2024optscaler}     & 3.2586 & 0.4912          & 6.5052          & 0.4693 & 1.3629 & 0.9598 & 3.7089 & 0.6401 \\
Llama3~\cite{touvron2023llama}        & 7.6830 & 1.1101          & 7.8901          & 0.6657 & 3.2022 & 1.5345 & 6.2584 & 1.1034 \\
TimeLLM~\cite{jintime}                & 3.1173 & 0.4423          & 5.3406          & 0.4513 & 1.4188 & 1.0059 & 3.2922 & 0.6332 \\
TimeMixer~\cite{wangtimemixer}        & 2.6458 & 0.4664          & 2.4406          & 0.3253 & 1.3918 & 0.9800 & 2.1594 & 0.5906 \\
iTransformer~\cite{liuitransformer}   & 2.3571 & 0.2800          & 2.3565          & 0.2875 & 1.3918 & 0.9804 & 2.0351 & 0.5160 \\
CCM+iTransformer (ours)                                     & 2.3677 & 0.2776          & 2.3206          & 0.2782 & 1.3608 & 0.9618 & 2.0164 & 0.5059 \\
TimesNet~\cite{wutimesnet}            & 2.2003 & \textbf{0.2609} & \textbf{2.2419} & 0.2602 & 1.3917 & 0.9791 & 1.9446 & 0.5001 \\
\textbf{CCM+TimesNet (ours)} &
  \textbf{2.1979 ↓0.11\%} &
  0.2612 ↑0.01\% &
  2.2503  ↑0.04\%&
  \textbf{0.2573 ↓1.09\%} &
  \textbf{1.3344 ↓4.12\%} &
  \textbf{0.9520 ↓2.77\%} &
  \textbf{1.9275 ↓0.88\%} &
  \textbf{0.4902 ↓2.00\%} \\ \hline
\end{tabular}%
}
\end{table*}

\end{appendix}
\end{document}